\documentclass[twocolumn,superscriptaddress]{revtex4-1}
\usepackage{amsmath, amsfonts, amssymb, graphicx, color, cancel, float, soul} 
\graphicspath{{./figures/}}

\newcommand{\firstequal}{These authors contributed equally.}
\newcommand{\lastequal}{These authors contributed equally.}

\newcommand{\E}{\mathbb{E}}
\newcommand{\equaltext}[1]{\ensuremath{\stackrel{\text{#1}}{=}}}

\begin{document}

\title{MINIMALIST: Mutual INformatIon Maximization for~Amortized \\ Likelihood Inference from~Sampled Trajectories }

\author{Giulio Isacchini}
\thanks{\firstequal}
\affiliation{
  Laboratoire de Physique de l'\'Ecole normale sup\'erieure,
  CNRS, PSL University, Sorbonne Universit\'e, and Universit\'e Paris Cit\'e, 75005 Paris, France}
\affiliation{
  Max Planck Institute for Dynamics  and Self-organization,  
  Am Fa\ss berg 17, 37077 G\"ottingen, Germany }
\author{Natanael Spisak}
\thanks{\firstequal}
\affiliation{
  Laboratoire de Physique  de l'\'Ecole normale sup\'erieure,
  CNRS, PSL University, Sorbonne Universit\'e, and Universit\'e Paris Cit\'e, 75005 Paris, France}
\author{Armita~Nourmohammad}
\thanks{\lastequal}
\affiliation{
  Max Planck Institute for Dynamics  and Self-organization,  
  Am Fa\ss berg 17, 37077 G\"ottingen, Germany }
\affiliation{
  Department of Physics, University of Washington,  
  3910 15th Ave Northeast, Seattle, WA 98195, USA}
\affiliation{
  Fred Hutchinson Cancer Research Center,  
  1100 Fairview ave N, Seattle, WA 98109, USA}
\author{Thierry Mora}
\thanks{\lastequal}
\affiliation{
  Laboratoire de Physique de l'\'Ecole normale sup\'erieure,
  CNRS, PSL University, Sorbonne Universit\'e, and Universit\'e Paris Cit\'e, 75005 Paris, France}
\author{Aleksandra M. Walczak}
\thanks{\lastequal}
\affiliation{
  Laboratoire de Physique de l'\'Ecole normale sup\'erieure,
  CNRS, PSL University, Sorbonne Universit\'e, and Universit\'e Paris Cit\'e, 75005 Paris, France}

\begin{abstract}
Simulation-based inference enables learning the parameters of a model even when its likelihood cannot be computed in practice. 
One class of methods uses data simulated with different parameters to infer models of the likelihood-to-evidence ratio, or equivalently the posterior function. 
{Here we frame the inference task as an estimation of an energy function parametrized with an artificial neural network.
We present an intuitive approach where the optimal model of the likelihood-to-evidence ratio is found by maximizing the likelihood of simulated data. 
Within this framework, the connection between the task of simulation-based inference and mutual information maximization is clear, and we show how several known methods of posterior estimation relate to alternative lower bounds to mutual information. 
These distinct objective functions aim at the same optimal energy form and therefore can be directly benchmarked. We compare their accuracy in the inference of model parameters, focusing on four dynamical systems that encompass common challenges in time series analysis: dynamics driven by multiplicative noise, nonlinear interactions, chaotic behavior, and high-dimensional parameter space.}
\end{abstract}

\maketitle

\section{Introduction}
Model-based Bayesian inference relies on knowing the probabilistic description of a process. Traditional methods rely on computing the likelihood of the observed data given the model parameters, in order to maximize or sample from the posterior. For many models, in particular with multiple interacting degrees of freedom or hidden variables, the likelihood function may be impractical to evaluate. 
In cases where drawing data from the generative process is possible, simulation-based inference techniques can be used as a powerful alternative approach for characterizing the underlying model. 

Population genetics provides many examples of such problems. The observed quantities in this context are often based on sequencing data and are  ``far'' from the quantities described by population dynamics models: it is in principle possible to write down likelihood functions, but they typically depend on a number of hidden variables that need to be marginalized out, making their evaluation impractical. 
Approximate Bayesian Computation (ABC) was first used for posterior inference in the context of population genetics~\cite{Tavare1997}, and since then numerous new approaches to simulation-based inference have been developed to answer particular questions of phylodynamics and sequencing data analysis. 

\pagebreak 

More broadly, methods for simulation-based inference can be organized in two classes~\cite{Cranmer2020, SBIBenchmark}. In the first class, observations and simulated data are compared within the inference process, {as in the original ABC approach~\cite{ABC, ABC3}}. Methods belonging to the second class proceed in two stages. First, they use a large number of simulations to learn an approximate model for the likelihood function~\cite{LP, LP2}, or alternatively the posterior function~\cite{PI}, or the likelihood{-to-evidence} ratio~\cite{Cranmer2015, LFIRE, Hermans2020}) that is amortized over the simulation examples. The amortized model is then used to evaluate the likelihood of observations and to evaluate the posterior for model parameters. With theoretical developments in machine learning and improvements in computing power, this class of algorithms has seen a renewed interest in the past years.

\begin{figure*}
\begin{center}
\includegraphics[width=.81\textwidth]{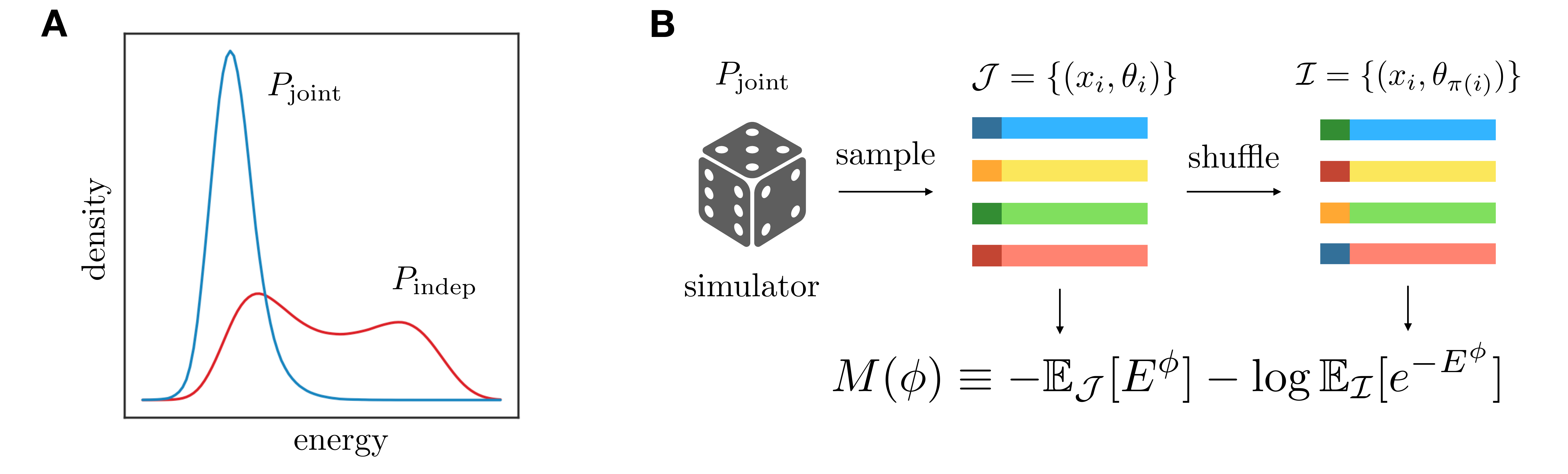}
\caption{{{(A) Distribution of energies of independent and joint pairs $(x,\theta)$.} Pairs from the joint distribution have lower energy $E$, while pairs from the independent distribution have higher energy, as the majority of these independent samples are relatively unlikely under a joint model. {(B) Schematic of the method.} We first sample parameters $\theta$ from the prior $P(\theta)$ and then sample observations $x$ from the simulator $P(x| \theta)$ to obtain pairs $(x_i, \theta_i)$. In order to generate pairs $(x_i,\theta_j)$ from the independent distribution we shuffle the two initial vectors. Both sets are used to infer a model of energy $E^\phi$ by maximizing a log-likelihood in \eqref{eq:MINE} with $\phi$ parameters of the artificial neural network used for the inference. }}
\label{illustration}  
\end{center}
\end{figure*}

{ Here, we model the likelihood-to-evidence ratio as a Boltzmann factor and use it to infer the corresponding energy function. } We show that the maximum-likelihood estimation of this factor is equivalent to the maximization of {a lower bound to} mutual information between parameters of the simulation and the simulated data.  {We exploit this equivalence by testing different lower bounds to mutual information as objective functions for simulation-based inference and optimize the parameters of artificial neural networks to approximate the posterior.}
We compare this approach to a recently developed method~\cite{LFIRE, Hermans2020},  where a model of the likelihood-to-evidence ratio is learned through the optimization of a binary classifier operating on simulated data {and which has been shown to provide state-of-the-art performance on a variety of tasks \cite{SBIBenchmark}}. We assess the accuracy {of the methods}  to infer the parameters of {four dynamical systems: the Ornstein-Uhlenbeck process, a multidimensional stochastic process with analytically tractable posterior; the birth-death process with multiplicative noise-driven dynamics; the Susceptible-Infected-Recovered (SIR) model of epidemiology, a simple system with elementary non-linear interactions; the Lorenz attractor, a dynamical system that exhibits chaotic behavior. 
The task of finding the parameters of stochastic processes by maximizing the likelihood of observed discrete trajectories is generically difficult \cite{Ferretti2020}. Our experiments encompass the most common challenges in the analysis of time-series data and together show that simulation-based inference offers a viable alternative to analytical methods.}

\section{Methods}

We aim to estimate parameters $\theta$ of a model given a set of data $x$ obtained from stochastic simulation of that model, $P(x|\theta)$, with a prior $P(\theta)$. 
First, we reinterpret the likelihood-to-evidence ratio in terms of a Boltzmann factor
\begin{equation}
\frac{P(x|\theta)}{P(x)}=\frac{P(x,\theta)}{P(x)P(\theta)}\equiv \frac{P_{\text{joint}}(x,\theta)}{P_{\text{indep}}(x,\theta)}=\frac{1}{Z}e^{-E(x,\theta)},
\label{ratio}
\end{equation}
where the energy function $E(x,\theta)$ captures how the joint distribution of data and parameters $P_{\text{joint}}(x,\theta)=P(x|\theta)P(\theta)$ deviates from the independent distribution $P_{\text{indep}}(x,\theta) = P(x)P(\theta)$, {as shown schematically in Fig.~1A,  in which the pairs $(x,\theta)$ sampled from $P_{\text{joint}}$ have lower energies than the samples from $P_{\text{indep}}$.}

The energy $E(x,\theta)$ is generally a non-linear function describing the dependence between data and parameters. $Z$ is the partition function, which ensures that the probability density $P(x,\theta)$ is normalized. {$E$ and $Z$ are each defined up to constants (additive for $E$, multiplicative for $Z$).}
Given the energy function, $E(x,\theta)$ we recover the posterior probability density $P(\theta|x) =\frac{1}{Z}e^{-E(x,\theta)} P(\theta)$.

We will now describe an inference scheme to learn a model of the energy function from simulated data, relying on the flexibility of artificial neural networks. Specifically, we approximate the energy $E$ by a multi-layered network $E^\phi$ characterized by a set of parameters $\phi$. Under a given model $E^\phi$, the joint distribution is approximated as:
\begin{equation}
P^{\phi}_{\text{joint}}(x,\theta)=\frac{1}{Z^{\phi}} e^{-E^{\phi}(x,\theta)} P_{\text{indep}}(x,\theta).
\label{phi-likelihood}
\end{equation}
We simulate samples from the joint distribution, denoted $\mathcal{J}=\{(x_i,\theta_i)\}_{i=1}^N$ by drawing a model parameter from a prior distribution,  $\theta_i \sim P(\theta)$  and simulating $x_i\sim P(x|\theta_i)$.

To learn the neural network parameters $\phi$, we need to maximize the log-likelihood of the simulated sample $\mathcal{J}$ under a given model $E^\phi$:
\begin{align}
\mathcal{L}(\phi;\mathcal{J}) &=N\E_\mathcal{J}[\log P^{\,\phi}_{\text{joint}}] \nonumber \\
                                              &=  N\left(-\E_\mathcal{J}[E^\phi]  - \log Z^\phi  + \E_\mathcal{J}[\log P_{\text{indep}}]\right).
\label{eq.likelihood}
\end{align}
where $\E_\mathcal{J}[\cdot]$ denotes the empirical average over samples $\mathcal{J}$. The partition function is approximated using importance sampling on samples drawn from $P_{\rm indep}$:
\begin{equation}
Z^\phi = \int e^{-E^\phi(x,\theta)}P_{\rm indep}(x,\theta)dx d\theta \approx \E_\mathcal{I}[e^{-E^\phi}],
\label{eq.Zphi}
\end{equation}
where $\E_\mathcal{I}[\cdot]$ is the counting measure over a large set $\mathcal{I}$ of independently drawn parameter and data pairs $(x,\theta)\sim P_{\rm indep}(x,\theta)=P(\theta)P(x)$. In practice, $\mathcal{I}$ may be obtained by shuffling the indices of $\mathcal{J}$ \cite{Hermans2020}, $\mathcal{I}=\{(x_i,\theta_{\pi(i)})\}$, where $\pi$ is a random permutation of $N$ elements, possibly multiple times.
Counting all possible combinations, the set $\mathcal{I}$ can have maximal size $\max(N_\mathcal{I})=N^2-N$ under a fixed simulation budget. We denote the relative size of the two sets by $k=N_\mathcal{I}/N$.

With this estimate of $Z^\phi$, and noting that the last term of eq.~\ref{eq.likelihood} does not depend on $\phi$, the problem is reduced to maximizing{
\begin{equation}
  M(\phi;\mathcal{I},\mathcal{J})\equiv-\E_\mathcal{J}[E^\phi]  - \log \E_\mathcal{I}[e^{-E^\phi}],
  \label{eq:MINE}
 \end{equation}
which in the infinite data limit constitutes a lower bound (the Donsker-Varadhan representation) to mutual information $I(X;\Theta)$ between simulated data and simulation parameters,
\begin{align}
M(\phi;\mathcal{I},\mathcal{J}) & \equaltext{$N\to\infty$} \int P_{\rm joint}(x,\theta)\log\frac{P_{\rm joint}^\phi(x,\theta)}{P_{\rm indep}(x,\theta)}\,dx\,d\theta& \nonumber \\ 
& = I(X;\Theta)-D_{\rm KL}(P_{\rm joint}^{\;\;}\Vert P_{\rm joint}^\phi) \nonumber \\
&\leq I(X;\Theta),
 \end{align}
where $D_{\rm KL}(P_{\rm joint}^{\;\;}\Vert P_{\rm joint}^\phi)$ is the Kullback-Leibler divergence between the true joint distribution and its model~(\ref{phi-likelihood}).} This bound was extensively studied in \cite{MINE} as an estimate of the mutual information from discrete samples drawn from joint distributions. Here, we will use this representation to learn the energy function $E^\phi$, which approximates the likelihood-to-evidence ratio. We refer to this method as Mutual Information Neural Estimation (MINE). {Its rationale is presented schematically in Fig.~1B.}

The connection to mutual information estimation established above opens the possibility of employing other empirical mutual information estimates to perform simulation-based inference. An alternative lower bound to $I(X;\Theta)$, first introduced in~\cite{FDIV}, is the so-called f-divergence representation (FDIV),
\begin{align}
L_f(\phi;\mathcal{I},\mathcal{J}) \equiv - \E_{\mathcal{J}}[E^\phi]  -\E_{\mathcal{I}}[e^{-E^\phi-1}].
\label{f-div}
\end{align}
This estimator defines an alternative objective function to~\eqref{eq:MINE} that can be used to infer an optimal energy model $E^\phi$. Note that in the limit of infinite data $N\to\infty$, and when the class of models $\{E^\phi\}_\phi$ can represent  {the true energy} exactly, the maxima of \eqref{eq:MINE} and \eqref{f-div} are both reached at the true value of $E(x,\theta)$ where they give the true value of $I(X;\Theta)$ {(see Appendix A)}. Outside of this limit, using one of these objective functions may prove more beneficial. In particular, the second term of $L_f(\phi;\mathcal{I},\mathcal{J})$ and its gradients may be reliably estimated by averaging over small batches, unlike the second term of {$M(\phi;\mathcal{I},\mathcal{J})$} because of the logarithm, giving FDIV an advantage for stochastic gradient descent algorithms. While the Donsker-Varadhan bound on the mutual information is tighter, i.e.
$L_f(\phi;\mathcal{I},\mathcal{J})\leq  M(\phi;\mathcal{I},\mathcal{J})$ holds for $N\to\infty$~\cite{MINE}, it is unclear whether it might produce a more reliable estimate of $E(x,\theta)$.

A third alternative is to use the original approach for the likelihood-to-evidence ratio estimation proposed in \cite{LFIRE} and~\cite{Hermans2020}. 
The energy $E(x,\theta)$ may be rewritten in terms of a classifier between the two  hypotheses of $(x,\theta)$ originating from the joint or independent distribution in a mixture $P_{\rm mix}=\frac{1}{k+1}P_{\rm joint}+\frac{k}{k+1} P_{\rm indep}$ (in \cite{Hermans2020} $k=1$). We define:
\begin{align}
  d(x,\theta)\equiv P({\rm joint}|x,\theta) &=\frac{P_{\rm joint}(x,\theta)}{P_{\rm joint}(x,\theta)+k P_{\rm indep}(x,\theta)} \nonumber \\
  &= \frac{1}{1+k Z e^{E(x,\theta)}}.  \label{eq:classifier}
\end{align}

The classifier is parametrized by a neural network, $d=d^\phi$ and is trained by minimizing the binary cross-entropy
\begin{equation}
S(\phi;\mathcal{I},\mathcal{J}) =  - \E_\mathcal{J}[\log{d^\phi}] - k \E_\mathcal{I}[\log{\left(1-d^\phi\right)}].
\label{cross-entropy}
\end{equation}

Similarly to objectives \eqref{eq:MINE} and \eqref{f-div}, in the $N\to\infty$ limit and when the class of $\{d^\phi\}_\phi$ models contains the true $d$, this cross-entropy is minimized at the true value $d(x,\theta)$. In \cite{CCMI} it was shown that it can also be used as an estimator of mutual information by computing the mean logarithm of the predicted likelihood-to-evidence ratio, $\E_{\mathcal{J}}[\log(k d^\phi/(1-d^\phi))]$. 
{Indeed, in the infinite data limit, all three objective functions, (\ref{eq:MINE}), (\ref{f-div}), and (\ref{cross-entropy}), share the same optimum; see Appendix~A.}
For high-dimensional random variables, binary cross-entropy (\ref{cross-entropy}) sets a tighter lower bound and a more accurate estimate of the mutual information than the f-divergence estimator (\ref{f-div}) \cite{CCMI}. However, its accuracy was not directly compared to the proposed estimator in eq.~\ref{eq:MINE}. We will refer to the inference approach based on minimizing the binary cross-entropy loss in eq.~\ref{cross-entropy} as BCE.

Finally, once an energy model $E^\phi$ has been trained by optimizing $M(\phi;\mathcal{I},\mathcal{J})$, $L_f(\phi;\mathcal{I},\mathcal{J})$, or $S(\phi;\mathcal{I},\mathcal{J})$ over $\phi$, the posterior of parameters given an observation $x$ may be calculated as $P(\theta|x)=(1/Z^\phi)e^{-E^\phi(x,\theta)}P(\theta)$. {We note that if the prior is changed, the energy function needs to be re-inferred.} 
When $\theta$ is of high dimension, scanning the posterior for all possible values of $\theta$ may be impractical.
In that case, we generate samples of $\theta$ from the posterior using a Markov-Chain Monte-Carlo method with Metropolis-Hasting acceptance probability:
\begin{equation}
{\rho(\theta \xrightarrow{} \theta') = \min \left(1, \frac{q(\theta'|\theta)P(\theta')}{q(\theta|\theta')P(\theta)} e^{-(E^\phi(x,\theta')-E^\phi(x,\theta))} \right),}
\end{equation}
with $q$ an ergodic Markov transition probability in the parameter space. This procedure generalizes in a straightforward way to the case of multiple observations drawn with the same set of parameters. 
 \section{Relations to other work}
 
The presented methods are related to several recent approaches. For completeness, we discuss the similarities and differences between the presented and other methods. 

\subsection*{Posterior inference methods}

An alternative to likelihood-to-evidence ratio estimation (RE) is the framework of Neural Posterior Estimation (NPE) which consists in fitting a conditional density estimator directly to the posterior. Several recently-developed NPE algorithms~\cite{SNPEA, SNPEB, SNPEC} have been compared with the BCE method in~\cite{SBIBenchmark}, as part of a public benchmark of ABC, NPE, and RE methods across several simulation-based inference tasks. The comparison reveals that there is no single best algorithm, NPE and RE approaches yield similar performance and consistently outperform ABC. The benchmark we present below is limited to different RE methods and compares alternative loss functions. 

\subsection*{Noise contrastive estimation}

{In~\cite{Contrastive}, the NPE  and the RE methods have been presented as two instances of a more general scheme, which the authors termed as {\em noise contrastive estimation}.}
The learning algorithm introduced in~\cite{Contrastive} is based on a multi-sample loss function that allows interpolating between the two approaches. Importantly, ref.~\cite{Contrastive} also suggests a link between the field of mutual information estimation and simulation-based inference since the multi-sample loss function is also a lower bound to $I(X;\Theta)$. Here we formalize this link and test its applicability on a variety of examples by proposing two new methods (MINE and FDIV) for likelihood-to-evidence ratio estimation. We detail how our approach fits within the noise contrastive estimation scheme in Appendix B.

\subsection*{Learning optimal experimental designs\\ and sufficient statistics}

An important direction of research in simulation-based inference aims at finding experimental designs that are most informative about the model parameters {\cite{Kleinegesse2020, Foster2019}}. In~\cite{Kleinegesse2019}, an optimal design is found by maximizing mutual information between design choice and target variable of interest in the experiment. This task extends the original RE framework to optimize over an additional hidden variable, the design, that specifies the simulation setup.

In~\cite{Chen2021a} mutual information is maximized to build sufficient summary statistics of the simulated data. This technique allows for automatically learning an optimal representation of the data without expert knowledge of process-specific observables. The statistics obtained this way can then generically be parsed as input to simulation-based inference methods.
While the procedure to find summary statistics is similar to the MINE and BCE methods described here,  the algorithm of~\cite{Chen2021a} does not exploit the energy model for posterior inference. Instead, the model is discarded and the statistics are independently utilized with another simulation-based inference method.  

Concurrently to the first version of this paper, a new development in this line of research~\cite{Kleinegesse2021} has been proposed. The authors compare different mutual information lower bounds for Bayesian experimental design and observe a consistent trend with our findings, namely that the BCE method performs overall better than other estimators. It is an independent validation of our results using a closely related task.

\section{Experiments}

We set out to examine the 3 presented methods for estimating the likelihood-to-evidence ratio (MINE, FDIV, and BCE) to infer the parameters of simple dynamical models from discrete samples of their trajectories. We chose 4 contexts that together encompass the range of difficulties in the inference of model parameters: (i) the stochastic birth-death process, (ii) the epidemiological Susceptible-Infected-Recovered (SIR) process, (iii) the multidimensional Ornstein-Uhlenbeck process, and (iv) the chaotic system of  Lorenz attractor. Example trajectories of each model are shown in Fig.~2.

\begin{figure*}
\begin{center}
\includegraphics[width=.9\textwidth]{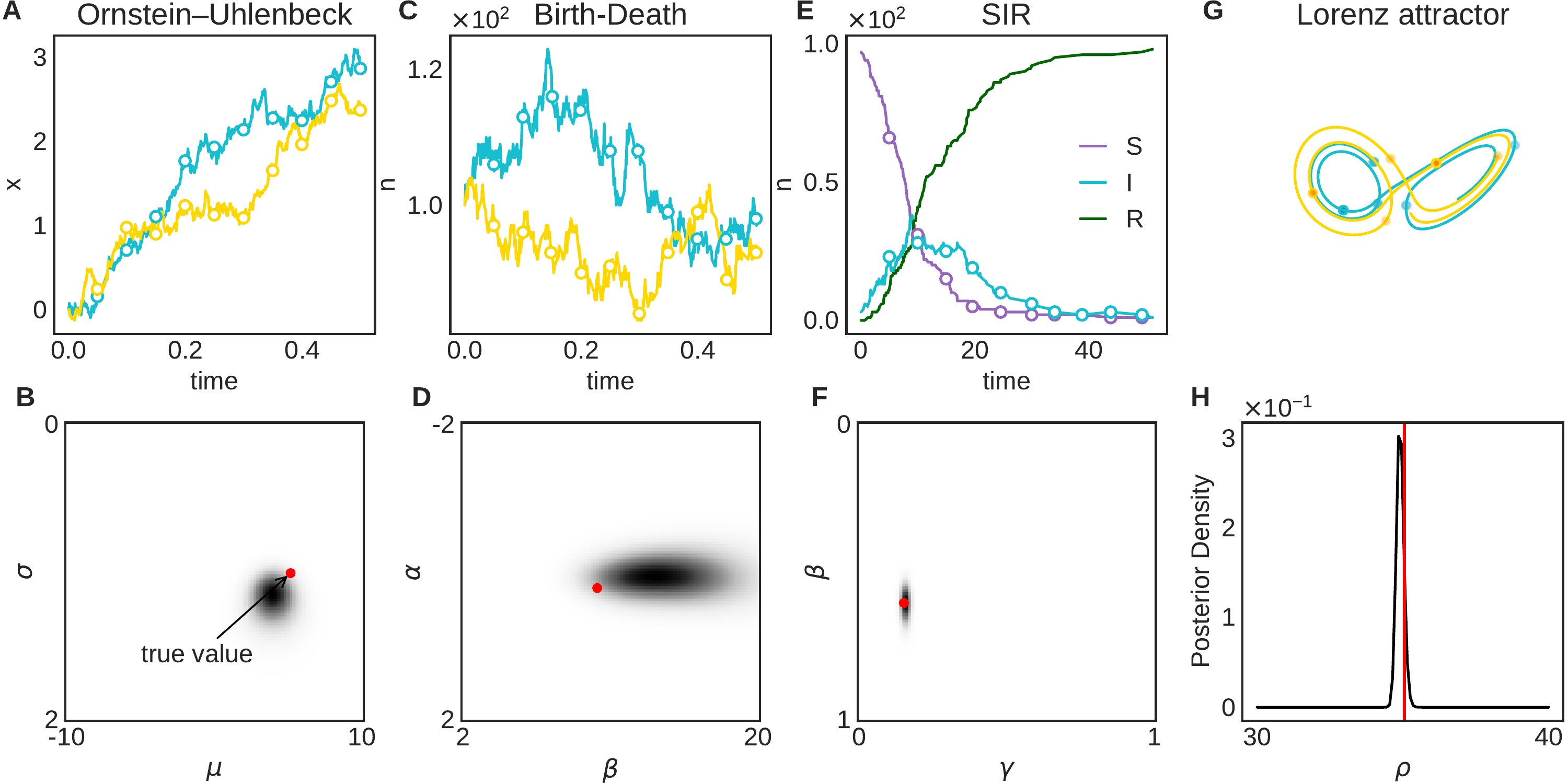}
\caption{Example trajectories (A,C,E,G) and posterior inference (B,D,F,H) for Ornstein-Uhlenbeck (A,B), Birth-Death (C,D), SIR (E,F) and Lorenz attractor (G,H). Trajectories are simulated with the parameter marked in red on the posterior plots. Circles indicate the discrete observations used for inference. Posteriors where estimated over 10 trajectories.}
\label{illustration}  
\end{center}
\end{figure*}

\subsection*{Ornstein-Uhlenbeck process}
The Ornstein-Uhlenbeck (OU) process is a multidimensional Markov process driven by additive Gaussian white noise. It is applied in many branches of science, notably to describe the velocity of a Brownian particle~\cite{Uhlenbeck1930}, the fluctuations of interest rates~\cite{Vasicek1977} or evolution of continuous phenotypic traits~\cite{Cavalli1967, Felsenstein1988}. The trajectories are a solution of a stochastic differential equation: 
\begin{equation}
dx= -\gamma \left(x - \mu \right) dt + \sqrt{2}\sigma dW,
\end{equation}
where $\mu$ is the stationary mean and $\gamma$ is the damping matrix, assumed to be symmetric. $W$ stands for the multidimensional Wiener process, and $\sigma$ is the noise amplitude. We use the Euler-Maruyama integration scheme to obtain the numerical solutions of this equation~\cite{springerBook}. 
The corresponding Fokker-Planck equation for this process can be solved to obtain the true posterior (see Appendix C).

In one dimension, we infer the mean $\mu$ and the noise strength $\sigma$, setting $\gamma=1$ and using uniform priors $P(\mu)=\mathcal{U}(-10,10)$ and $P(\sigma)=\mathcal{U}(0,2)$. 

To study how { the performance of each method} scales with dimension, for $1\le d \le 5$, we fix $\mu=0$, $\sigma = \mathbb{I}$, where $\mathbb{I}$ is the identity matrix, and infer the damping matrix parametrized as $\gamma = \mathbb{I} + \epsilon g$, where $g$ is a Gaussian orthogonal matrix and $\epsilon<1$, which  ensures that the damping matrix is positive definite (see Appendix C for more details). Since $\gamma$ is symmetric, in the $d$-dimensional case we have ${d \choose 2}$ parameters to infer. The prior is given by the Gaussian Orthogonal Ensemble distribution density, $P(g)\propto \exp{(-d\,\mathrm{Tr}(g^2)/4)}$.

\subsection*{Birth-death process}
The birth-death process is a discrete one-dimensional Markov process with multiplicative demographic noise. The number of individuals $n$ is subject to variation due to stochastic birth and death events occurring at rates $n\lambda$ and $n\delta$, respectively,
\begin{equation}
n \xrightarrow{n\lambda} n+1, \qquad n \xrightarrow{n\delta} n-1.
\label{birth-death}
\end{equation}
We use the Gillespie algorithm to sample trajectories from this process~\cite{Gillespie1977}. We parametrize the process with the average exponential drift $\alpha=\lambda-\delta$, and the noise timescale $\beta= \lambda + \delta$. We use uniform priors for both of these variables: $P(\alpha)=\mathcal{U}(-2,2)$ and  $P(\beta)=\mathcal{U}(2,20)$. 

\subsection*{SIR model} 
The Susceptible-Infected-Recovered (SIR) model is a staple of epidemiological modeling. Any member of {the }susceptible population $S$ can be infected at rate $\beta$ upon contact with one of $I$ infected individuals. The infected individuals can become resistant $R$ at a rate $\gamma$:
\begin{equation}
S + I \xrightarrow{\beta} 2I, \qquad I \xrightarrow{\gamma} R.
\end{equation}
We simulate the trajectories of the SIR model using the Gillespie algorithm~\cite{Gillespie1977}. We infer the rates $\beta$ and $\gamma$ under uniform priors $P(\beta)=P(\gamma) = \mathcal{U}(0,1)$ given samples from the $(S,I)$ trajectories.

\subsection*{Lorenz attractor}
The Lorenz system is a 3-dimensional chaotic system governed by the equations, 
 \begin{equation}
\dot{x}=\sigma(y - x), \quad \dot{y}= x(\rho - z) - y, \quad \dot{z}= xy -\beta z.
\end{equation}
We simulate this deterministic process starting from a random position $(x_0+\eta, y_0,z_0)$, where $\eta$ is the noise in the initial position drawn from a uniform distribution, $\eta\sim\mathcal{U}(-0.1,0.1)$. We fix the parameters $\sigma=10$ and $\beta=8/3$ and set out to infer $\rho$. The ensemble of trajectories starting in the vicinity of $x_0$  diverge with the characteristic time set by the inverse of the largest Lyapunov exponent of the system $\lambda = \lambda(\rho)$. We start sampling from the trajectories at a random initial time drawn from a Gamma distribution, $t_0\sim \Gamma(k=5,\theta=2)$. We then take $5$ samples from each trajectory at time windows that are larger than the characteristic time for chaotic divergence $\Delta t=2\lambda^{-1}$. We set $\lambda\simeq 0.905$, which is the Lyapunov exponent for $\rho=28$, a transition point where some but not all the solutions of the Lorenz system are chaotic. We infer the parameter $\rho$ in a chaotic regime using a uniform prior $P(\rho) = \mathcal{U}(30,40)$. 

\begin{figure*}
\begin{center}
\includegraphics[width=.9\textwidth]{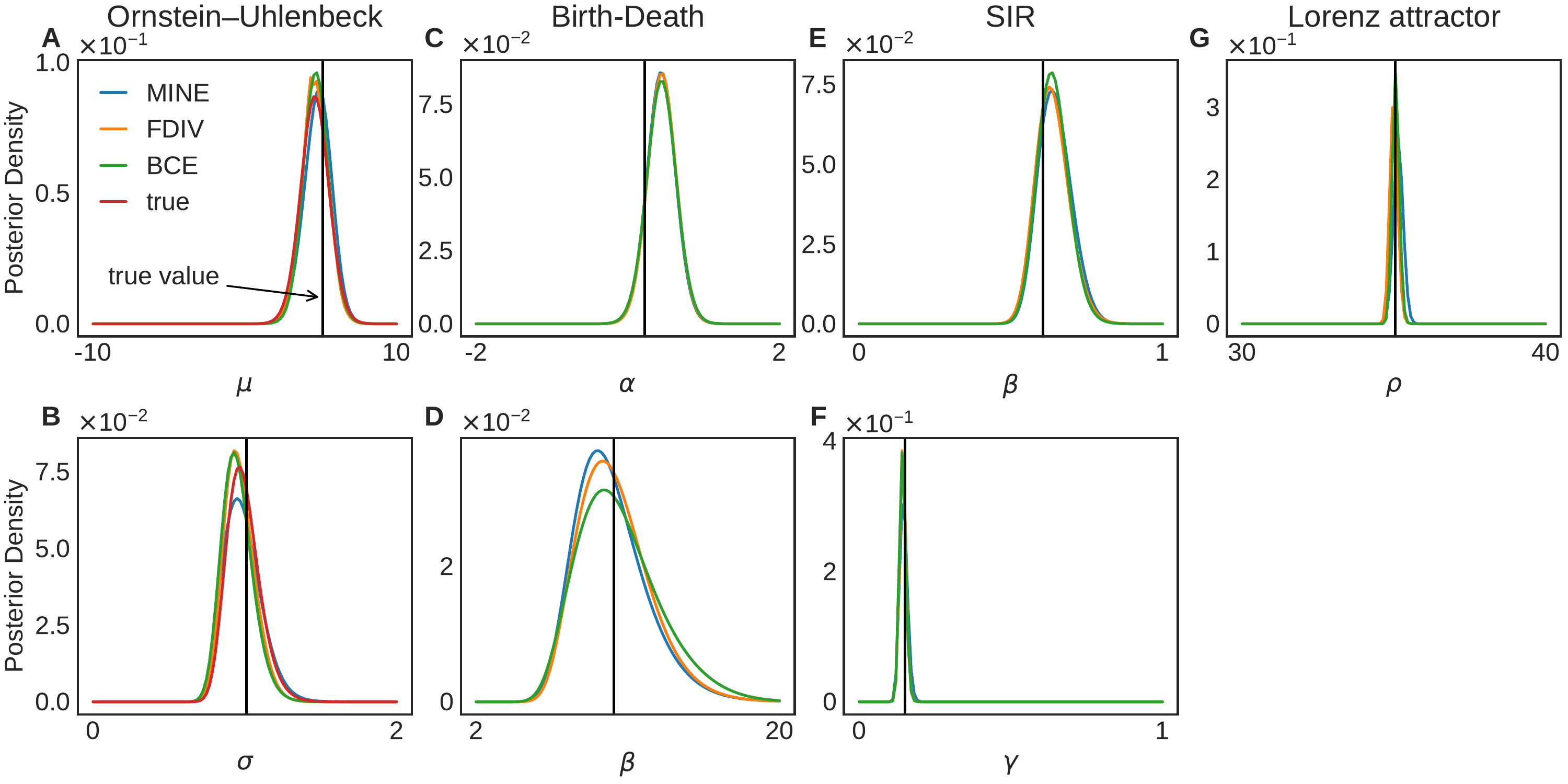}
\caption{{Convergence of the posteriors} for Ornstein-Uhlenbeck (A,B), Birth-Death (C,D), SIR (E,F) and Lorenz attractor (G). In order to evaluate the posteriors we first choose a reference hypothesis $\theta^*$: for Ornstein-Uhlenbeck $\mu^*=5$ and $\sigma^*=1$, for Birth-Death $\beta^*=10$ and $\alpha^*=0.2$, for SIR $\beta^*=0.6$ and $\gamma^*=0.2$, for Lorenz attractor $\rho^*=35$. We show posteriors $P^\infty(\theta|x_{1:M})$ calculated using models trained with $N=10^7$ trajectories. For Ornstein-Uhlenbeck process the exact posterior density $P(\theta|x_{1:M})$ is also shown.}
\label{illustration}  
\end{center}
\end{figure*}

The artificial neural networks used for all 3 methods were multilayer perceptrons~\cite{Perceptron0} with two hidden layers and a hyperbolic tangent activation function. This architecture choice was found to be expressive enough across tasks and, thanks to its simplicity, we could perform a well-grounded comparison of the three methods without advanced regularisation techniques (see Appendix D for details on hyperparameters choices). The methods were implemented using Tensorflow~\cite{tensorflow,keras} with extensive use of Numpy~\cite{harris2020array} and Scipy~\cite{scipy} libraries.

\section{Results}

Given enough data and a powerful enough neural network, we expect the optima of the objective functions $I$, $L_f$, and $S$ to converge, and the estimated energy function should approach the true value. 

{To confirm the validity of the proposed methods, in the first set of experiments we use a large number of simulations to study the convergence of the posterior functions. To this end, we choose a hypothesis $\theta^*$ and simulate $M$ trajectories, $x_{1:M}=\{x_m\}_{m=1}^M$, $x_m\sim P(x|\theta^*)$ with $M=2$ for SIR, and $M=5$ for the other tasks.} We evaluate the posteriors
\begin{equation}
\hat{P_l}(\theta|x_{1:M})=\frac{1}{(Z^\phi_l)^M}\exp\left[-\sum_{m=1}^ME^\phi_l(x_m,\theta)\right]P(\theta),
\end{equation}
with $l$ indexing one of the three methods (MINE, FDIV, or BCE) and $\phi$ is the optimal one for each method (we dropped the explicit dependency on $\phi$ in $\hat P_l$ for ease of notation). The posteriors converge when the amortized inference is done on a training set with at least $N=10^7$ samples (and $10^7$ samples for validation); see Fig.~3. We define a reference posterior $P^\infty(\theta|x_{1:M}) \equiv \langle \hat{P_l}(\theta|x_{1:M})\rangle_l$, obtained with $N=10^7$ as the average over three estimators.
In the case of Ornstein-Uhlenbeck process, where the true posterior $P(\theta|x_{1:M})$ can be calculated analytically,  $P^\infty(\theta|x_{1:M})$ agrees with the analytical prediction. {This first result confirms the validity of our approach.}

\begin{figure*}
\begin{center}
\includegraphics[width=.9\textwidth]{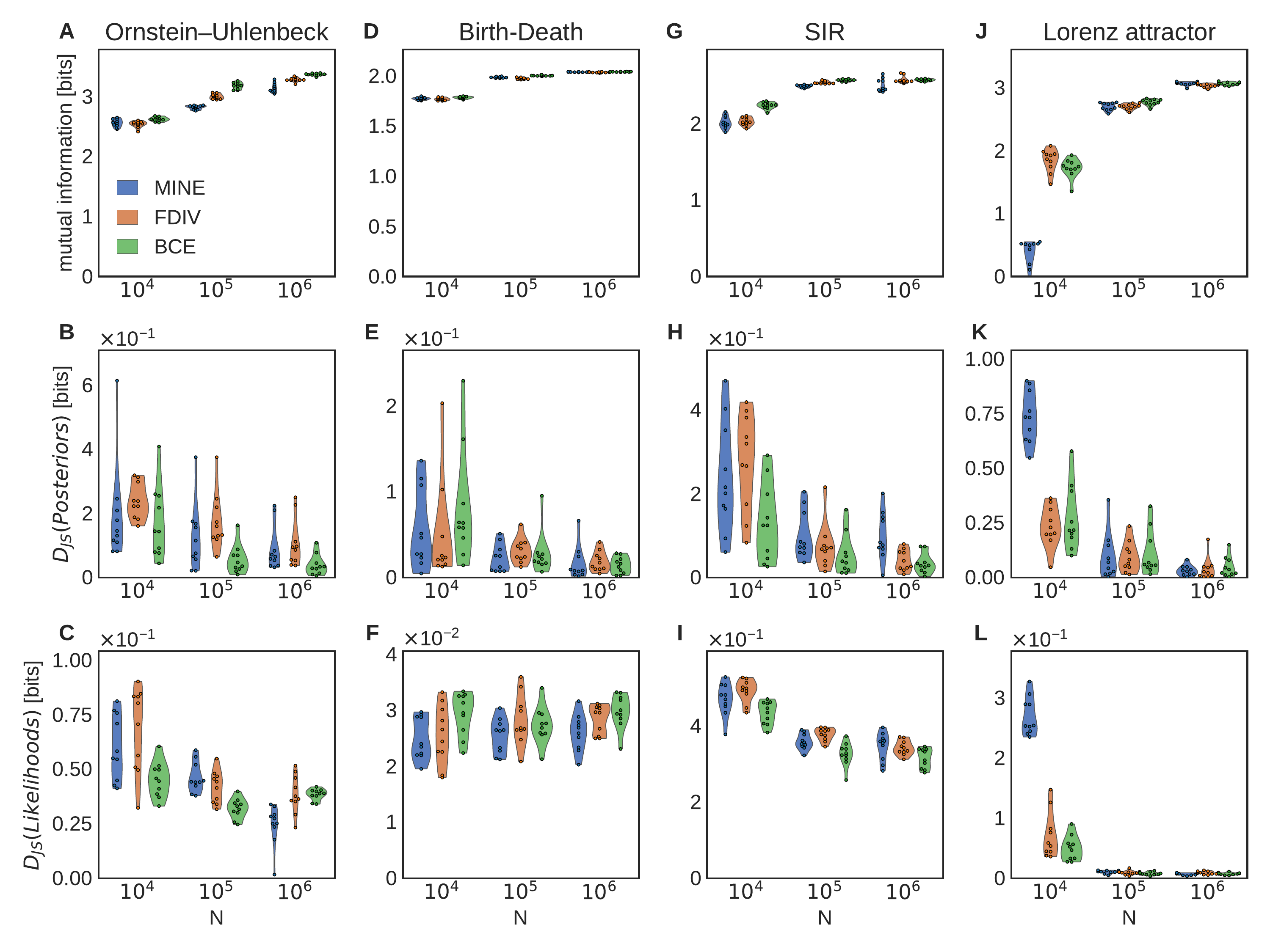}
\caption{{Benchmarking of the methods.} 
We compare the three objectives $M$, eq. \ref{eq:MINE} (MINE), $L_f$, eq. \ref{f-div} (FDIV), and $S$, eq. \ref{eq:classifier} (BCE) for 3 different metrics. We perform 10 replicates of the inference and comparison for simulation budgets $N\in \{10^4, 10^5, 10^6\}$ for 4 systems: Ornstein-Uhlenbeck (A,B,C), Birth-Death (D,E,F), SIR (G,H,I) and Lorenz attractor (J,K,L). In the first row we compare the mutual information on held out test data using  eq. \ref{eq:MINE} with the estimated $E^\phi$. For the following two metrics we need to instead choose an hypothesis $\theta^*$, see Fig 3. for the exact values. In the second row we compare the Jensen-Shannon divergence $D_{\rm JS}(P^\infty(\theta|{x_{1:M}}), \hat{P}^N_l(\theta|{x_{1:M}}))$ between the reference and inferred posteriors. In the last row we compare the Jensen-Shannon divergence $D_{\rm JS}(P(x|\theta^*), \hat{P}^N_l(x|\theta^*))$ using sampled trajectories from the simulator $P(x|\theta^*)$ and the inferred distribution $\hat{P_l}(x| \theta^*)$.}
\label{illustration}  
\end{center}
\end{figure*}

With reducing sample size $N$, the amortized posteriors differ. To study the performance of the three methods under different simulation budgets $N$ for each task we simulate $N_{\rm tot}=2 \times 10^7$ samples $\mathcal{J}=\{(x_i,\theta_i)\}$.  
We perform the inference of the amortized likelihood-to-evidence ratio with varying simulation budgets, where {both the training and the validation data} are equal-sized subsamples of $\mathcal{J}$ with $N\in \{10^4, 10^5,10^6\}$. Inference and comparison are performed 10 times on independent subsamples of $\mathcal{J}$. To obtain samples from the independent set $\mathcal{I}$ we shuffle the joint samples $k=5$ times, and so $N_{\mathcal{I}}=5N$. A larger shuffled data $N_{\mathcal{I}}$ can improve the inference but at the cost of computing power, which sets a trade-off between performance and training time.

We compare the accuracy of the 3 inference methods (MINE, FDIV, BCE) for the 4 tasks (OU, Birth-death, SIR, Lorenz) based on the three following metrics.
\begin{figure*}
\begin{center}
\includegraphics[width=1.\textwidth]{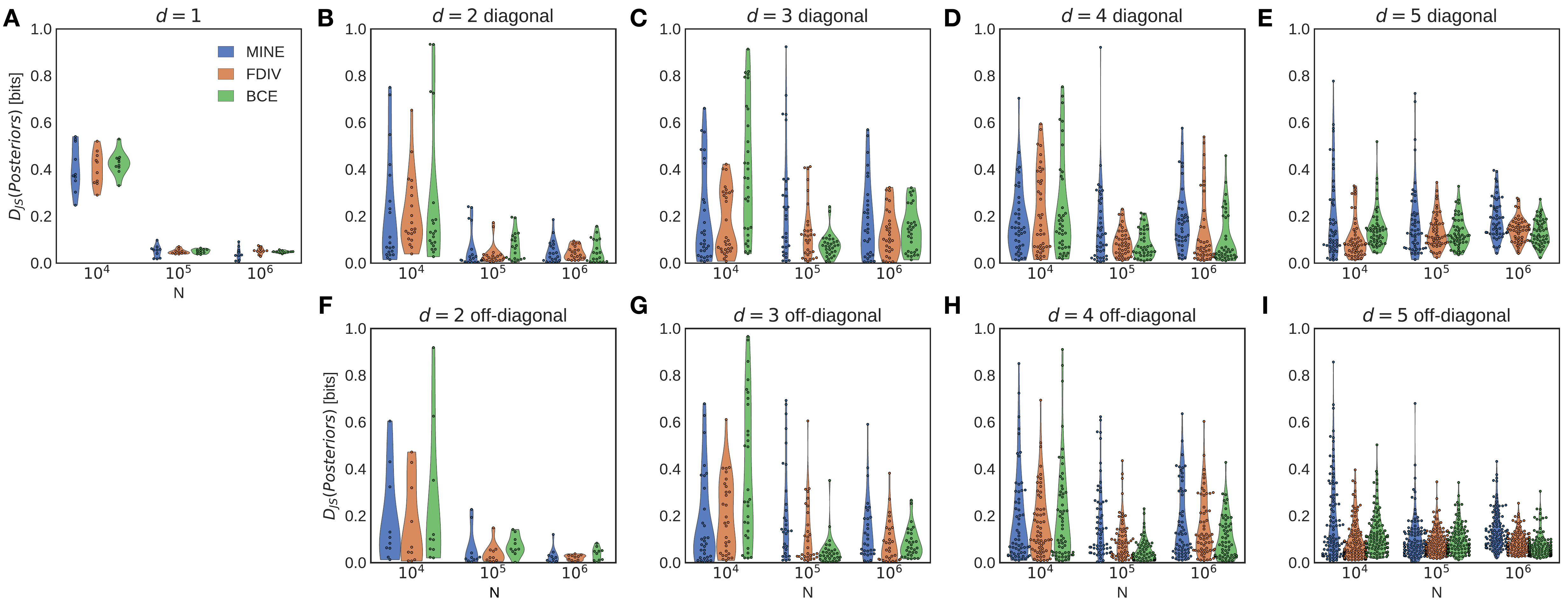}
\caption{{The Ornstein-Uhlenbeck process in $d\ge1$.}
We compare the three objective functions $M$ (MINE), $L_f$ (FDIV), and $S$ (BCE) for dimension $d=1,2,3,4,5$. We perform 10 replicates of the inference with simulation budgets $N = 10^4,10^5,10^6$ for changing dimension $d=1,2,3,4,5$. We compare the posterior with the analytical prediction for a hypothesis $(g^*)_{ij}=-1$. We compute the Jensen-Shannon divergence $D_{\rm JS}(P(\theta|{x_{1:M}}), \hat{P}^N_l(\theta|{x_{1:M}}))$ between the true and inferred posteriors for each element of the damping matrix $\gamma$ independently. We show the divergence for diagonal (A-E) and off-diagonal terms (F-I).}
\label{illustration}  
\end{center}
\end{figure*}

\subsection*{Global comparison} 
The first metric used for the benchmark is the mutual information given a density estimator, computed with eq.~\ref{eq:MINE}. For each $N$, it is evaluated on test data composed of the remaining $N_{\rm tot} -N$ samples. Unlike the other two comparisons (see below), it is a global metric that tests the approximation of the likelihood-to-evidence ratio over all $\theta$ and $x$ values. 
For this reason, we use it to perform hyperparameter tuning for each task and each objective with $N=10^5$, see Appendix D.

For all four tasks, the value of the estimated mutual information grows with the simulation budget and yields comparable performances for the 3 methods (Fig.~4 A, D, G, J). For the Ornstein-Uhlenbeck process, the BCE method reaches consistently higher values of mutual information. For the Lorenz attractor, the MINE method is significantly outperformed by the other methods in the low data limit ($N=10^4$).

\subsection*{Posterior comparison}  
The objective of simulation-based inference is to find the posterior distribution over model parameters. To characterize the inference accuracy as a function of the simulation budget $N$ and the method $l$,  we evaluate the Jensen-Shannon divergence between the inferred and the reference posterior $D_{\rm JS}(P^\infty(\theta|{x_{1:M}}), \hat{P}^N_l(\theta|{x_{1:M}}))$ (where $D_{\rm JS}(p,q)=(1/2)\int [p(x)\log(p(x)/m(x))+q(x)\log(q(x)/m(x))]dx$ with $m(x)=(p(x)+q(x))/2$), by scanning through the parameter space with the prior $P(\theta)$.
A larger Jensen-Shannon divergence indicates a larger deviation between the inferred posterior and the reference (i.e., a lower performance). 

All methods show comparable performances, and the Jensen--Shannon divergence decays as a function of the simulation budget $N$ (Fig.~4 B, E, H, K, and Fig.~5 for the Ornstein-Uhlenbeck process with $d\ge1$). {At $N=10^4$ the accuracy of the posterior inference is decreased for all objective functions, as reflected by the large variance of the $D_{\rm JS}$. In the case of the Lorenz attractor, this simulation budget is also insufficient for the MINE method which performs significantly worse than the classifier-based (BCE) and f-divergence (FDIV) estimators.}

Fig.~5 presents the comparison of the objective functions for the inference of the damping matrix elements in the high-dimensional Ornstein-Uhlenbeck process. We compare the performance in terms of the estimated posterior's divergence from the analytical prediction. We compute the Jensen-Shannon divergence independently for the marginal posterior distributions over each element of the damping matrix (a total of ${d\choose 2}$ unique elements in dimension $d$). We present the results independently for diagonal and off-diagonal elements of the matrix $\gamma$ (in dimension $d$ there are $d$ diagonal and $d(d-1)/2$ off-diagonal elements).

{The efficacy of the three methods is comparable and the average performance does not significantly decrease up to $d=5$, for which we estimate 15 elements of matrix $g$. At the same time, the variation in the Jensen-Shannon divergence increases with dimension as the inference task becomes more difficult. This is particularly pronounced for the MINE method in dimensions 4 and 5 at low ($N=10^4$) as well as intermediate ($N=10^5$) simulation budgets. In these instances, using the f-divergence objective function yields the best performance.}

\subsection*{Likelihood comparison}
The third metric is the  Jensen-Shannon divergence $D_{\rm JS}(P(x|\theta^*), \hat{P}^N_l(x|\theta^*))$ between the true and approximated likelihood for a given model $\theta$.
This $D_{\rm JS}$ cannot be directly evaluated by summing over $x$, because it is typically of high dimension. We thus rely on samples from these two distributions and infer an additional classifier to estimate $D_{\rm JS}$; see Appendix E. 

The performance of the 3 methods is comparable (Fig.~4 C, F, I, L). For the Ornstein-Uhlenbeck process, the BCE infers more accurate likelihood functions at $N=10^4$ and $N=10^5$ but it is outperformed by MINE at higher simulation budgets. 

The results of the benchmark shown in Fig~4 suggest that all estimators show reliable performances across different tasks and simulation budgets.  While the first metric is global and the two other metrics are local, they draw a consistent picture.  A higher simulation budget enhances the performance of all methods. The BCE method tends to perform better at the lowest simulation budget. All three methods perform similarly in the middle and high data regimes. 

\section{Conclusion}

We analyzed the problem of inferring an amortized estimator for the likelihood-to-evidence ratio over model parameters, using simulated data. We showed that this inference can be performed by maximization of the mutual information between simulated data and parameters of the model. This formulation captures an intuition that inference can be performed when we can extract the dependence between parameters and observed data, as measured by the mutual information. 
Our formalism opens up possibilities for using algorithms and techniques developed in the context of mutual information estimation~\cite{MIBounds} for inverse problems.

The likelihood function we propose is equivalent to the mutual information bound analyzed in~\cite{MINE}. However, while in~\cite{MINE} the focus is on the estimation of the absolute value of this quantity, we are interested in the inferred energy function that can be used to evaluate the posterior distribution for model parameters.  Previous work that used classifiers for simulation-based inference~\cite{LFIRE, Hermans2020} also fits naturally within our framework since logistic regression is linked to mutual information estimation~\cite{CCMI}.  The methods we studied rely on two lower bound estimators of mutual information, which are based on (i) the Donsker-Varadhan~\cite{MINE}, and (ii) f-divergence representations of the Kullback-Leibler divergence~\cite{FDIV}. It would be interesting to explore other known mutual information estimators for simulation-based inference~\cite{MIBounds}.

We showed that the mutual information-based methods (MINE and FDIV),  implemented in flexible neural networks, can reliably infer the posterior of the parameters and give consistent results with the previously proposed classifier-based technique (BCE)~\cite{LFIRE, Hermans2020} when the simulation budget is sufficient. We benchmarked the three approaches and found that their performances are comparable in the intermediate data regime, while in the low data regime the classifier-based method performs consistently better. The main limitation of the two proposed objective functions $M$ and $L_f$ is that they require large simulation budgets for accurate inference.

Our choice to implement the neural network as a multilayer perceptron with two hidden layers was motivated by having a simple and reliable architecture to better focus on the relative performance of the different objective functions. For the specific task of inference of model parameters from discrete samples of trajectories, absolute performance could be increased by choosing network architectures adapted to the data structure such as convolutional and recurrent layers.

Existing approaches to simulation-based inference, such as ABC, suffer from the need to define ad-hoc summary statistics to be matched between data and model. An important property of mutual information is its invariance upon the reparametrization of its variables. This enables inference and comparison of different parametrizations of the observed data, as different choices can be evaluated using the absolute value of the mutual information. 
A specific application that could be interesting to explore is inference for population genetics models, where the choice of summary statistics to use for ABC analysis has always {been} critical, and the ability to flexibly compare different parametrization choices greatly improves performance, as shown in~\cite{Chen2021a}. 
Another possibility would be to explore more principled regularization techniques such as the information bottleneck method~\cite{IBTishby}. This approach could be used to infer summary statistics of the data that are maximally informative of the parameters of the model. Then the summary statistics could be added as additional variables for the observations of related tasks, such as model extensions, in a transfer learning fashion.

In conclusion, our work helps to clarify the link between mutual information estimation and simulation-based inference. We believe that this connection can be a fruitful source of improved methods for amortized inference.

\subsection*{Code availability} The code for the algorithms presented in this paper is available at \url{github.com/statbiophys/MINIMALIST}

\section*{Acknowledgements}
This work was supported by the DFG grant (SFB1310) for Predictability in Evolution (AN, AMW, GI), the MPRG funding through the Max Planck Society (AN,  GI), the Royalty Research Fund from the University of Washington (AN), European Research Council grant, ERCCOG n. 724208 (AMW, TM, GI, NS), NSF CAREER Award, grant No: 2045054 (AN), and NIH MIRA award 1R35GM142795 - 01 (AN).

\bibliographystyle{unsrt}

\onecolumngrid

\newpage

\section*{Appendix}

\renewcommand{\theequation}{A\arabic{equation}}
\renewcommand{\thefigure}{A\arabic{figure}}
\setcounter{figure}{0}

\subsection{Different objective functions share an optimum} 
We study the following objective funtions:
\begin{eqnarray}
M(\phi;\mathcal{I},\mathcal{J})  &=& -\E_\mathcal{J}[E^\phi]  - \log \E_\mathcal{I}[e^{-E^\phi}],
\\
L_f(\phi;\mathcal{I},\mathcal{J}) &=& - \E_{\mathcal{J}}[E^\phi]  -\E_{\mathcal{I}}[e^{-E^\phi-1}],
\\
S(\phi;\mathcal{I},\mathcal{J}) &=&  - \E_\mathcal{J}[\log{d^\phi}] - k \E_\mathcal{I}[\log{\left(1-d^\phi\right)}]
\\ 
&=&  - \E_\mathcal{J} \left[\log{\frac{1}{1+kZe^{E^\phi}}}\right] -k \E_\mathcal{I} \left[\log{\frac{k}{k+Z^{-1}e^{-E^\phi}}}\right].
\end{eqnarray}
In the infinite data limit, the empirical averages converge and we can rewrite all objectives as functional of the energy model:
\begin{eqnarray}
M(E^\phi)  &=& - \int  E^\phi (x,\theta) P_{\rm joint}(x,\theta) \,dx\,d\theta - \log \int e^{-E^\phi (x,\theta)} P_{\rm indep}(x,\theta) \,dx\,d\theta,
\\
L_f(E^\phi) &=& - \int  \left(E^\phi (x,\theta) P_{\rm joint}(x,\theta) + e^{-E^\phi (x,\theta)-1} P_{\rm indep}(x,\theta) \right) \,dx\,d\theta,
\\
S(E^\phi) &=&  - \int  \Big(\log{\frac{1}{1+kZe^{E^\phi (x,\theta)}}} P_{\rm joint}(x,\theta) 
\\
                  &&  \;\;\;\;\;\;\;\;\;\;  + k \log{\frac{k}{k+Z^{-1}e^{-E^\phi (x,\theta)}}}P_{\rm indep}(x,\theta) \Big)dxd\theta.
\end{eqnarray}
In this limit the 3 optima of the objective functions are equivalent and recover the likelihood-to-evidence ratio. To see this, we take the functional derivative with respect to the energy model $E^\phi$, 
\begin{eqnarray}
\frac{\delta M}{\delta E^\phi(x,\theta)}  &=& -P_{\rm joint}(x,\theta) + \frac{1}{\int e^{-E^\phi(x,\theta)} P_{\rm indep}(x,\theta) \,dx\,d\theta} e^{-E^\phi(x,\theta)} P_{\rm indep}(x,\theta),
\\
\frac{\delta L_f}{\delta E^\phi(x,\theta)} &=& -P_{\rm joint}(x,\theta) + e^{-E^\phi(x,\theta)-1} P_{\rm indep}(x,\theta) ,
\\
\frac{\delta S}{\delta E^\phi(x,\theta)}  &=& k\left( \frac{Ze^{E^\phi(x,\theta)}P_{\rm joint}(x,\theta)-P_{\rm indep}(x,\theta)}{1+kZe^{E^\phi(x,\theta)}} \right) \left(1+\frac{\delta \log Z}{\delta E^\phi(x,\theta)}\right),\;\;\;\;\;
\end{eqnarray}
and we find they vanish at energies $E_M$, $E_f$ and $E_S$ respectively:
\begin{eqnarray}
E_M  &=& -\log{\frac{P_{\rm joint}(x,\theta) }{P_{\rm indep}(x,\theta) }} - \log{Z},
\\
E_f  &=& -\log{\frac{P_{\rm joint}(x,\theta) }{P_{\rm indep}(x,\theta) }} - 1,
\\
E_S &=& -\log{\frac{P_{\rm joint}(x,\theta) }{P_{\rm indep}(x,\theta) }} - \log{Z}.
\end{eqnarray}
All three are equal to the logarithm of the likelihood-to-evidence ratio up to constant factors. We note that the second derivatives are different in the 3 cases and therefore convergence to the optima $E_M$, $E_f$, and $E_S$ will in general be different.

\subsection{Noise Contrastive Estimation and mutual information} 

In this section, we show how our work fits within the framework of Noise Contrastive Estimation (NCE) and how it relates to the existing contrastive learning approaches to simulation-based inference.
The NCE methods estimate a probability density $p(y)$ by comparison to a reference noise distribution $q(y)$~\cite{Contrastive1,Contrastive2} : 
\begin{equation}
 p(y)=\frac{1}{Z}e^{-E(y)}q(y),
 \end{equation}
which reduces the problem to approximating the density ratio. The original method~\cite{Contrastive1} consists of the inference of the density ratio model using logistic regression (minimizing binary cross entropy) on samples from both distributions, $p,q$. This framework encompasses the likelihood-to-evidence ratio inference problem where $p(y)=P_{\text{joint}}(x,\theta)$ and $q(y)=P_{\text{indep}}(x,\theta)$ and one minimizes $S(\phi;\mathcal{I},\mathcal{J})$ to find $E^\phi$.

An alternative approach proposed in the Noise Contrastive Estimation literature ~\cite{Ranking, CollinsNoise} focuses on the estimation of conditional probability functions
 \begin{equation}
 p(y|z)=\frac{1}{Z(z)}e^{-E(z,y)}q(y),
\end{equation}
where now the partition function explicitly depends on the conditioned variable $z$. The new density ratio can be inferred by optimizing the the so-called ranking objective~\cite{Ranking}. This objective function is typically used to rank a positive sample from the target distribution $p(y|z)$ above $k$ samples from the reference noise $q(y)$ for the input $z$~\cite{CollinsNoise}.

In simulation-based inference, this family of methods has been used for posterior estimation, where  $p(y|z)=P(\theta|x)$ is the unknown posterior and $q(y)=P(\theta)$ is the prior. In our notations, the ranking objective function reads
\begin{equation}
L_r(\phi; \mathcal{J})=\E_{\mathcal{J}} \left[ \int \log\left( \frac{e^{-E^\phi(x,\theta)}}{e^{-E^\phi(x,\theta)}+\sum_{j=1}^k e^{-E^\phi(x,\theta_j)}}\right) \prod_{i=1}^k P(\theta_i)  d\theta_i \right].
\end{equation}
This method is known as the Sequential Neural Posterior Estimation (SNPE) proposed in~\cite{SNPEC}, building on the work in Refs.~\cite{SNPEA, SNPEB}. 
It's useful to note that the ranking loss $L_r(\phi; \mathcal{J})$ has also been used to construct a high-bias and low-variance estimator of mutual information~\cite{Oord1, Contrastive}. 

Ref.~\cite{Contrastive} proposes that also the binary classification approach of~\cite{Hermans2020} is a special case of the above inference for $k=1$. However, the ranking objective $L_r(\phi; \mathcal{J})$ with $k=1$ is distinct from $S(\phi;\mathcal{I},\mathcal{J})$ and the two methods cannot be identified as one.
In Ref.~\cite{CollinsNoise} the cross-entropy has been compared to the ranking loss and shown to generically outperform it in the context of Neural Language Processing.

\subsection{The Ornstein-Uhlenbeck process in dimension $d$} 
The trajectories $x(t)$ are solutions to a stochastic differential equation 
\begin{equation}
dx= -\gamma \left(x - \mu \right) dt + \sqrt{2}\sigma dW,
\end{equation}
where $x$ is a $d$-dimensional coordinate, $\mu$ its long-term average, $\gamma$ is a $d\times d$ damping matrix, $\sigma$ is the noise strength, and $W$ is a $d$-dimensional Wiener process. From a trajectory $x(t)$ we sample $n$ values every $\Delta t$ so that $x=\left\{x_i=x(i\Delta t)\right\}$. To find the analytical expression for the likelihood of these observations we write the corresponding Fokker-Planck equation for the density $P=P(x_i,t+\Delta t|x_{i-1},t)$,
\begin{equation}
\frac{dP}{dt} = - \nabla \left[ \gamma \left(x - \mu \right) P\right] + \sigma \sigma^T \nabla^2 P,
\end{equation}
solved with a multivariate Gaussian distribution density
\begin{equation}
P(x_i,t+\Delta t|x_{i-1},t) = \frac{1}{\sqrt{(2\pi)^d \det\Sigma}} e^{-\frac{1}{2}\left(x_i-\langle x_i\rangle\right)^T\Sigma^{-1}\left(x_i-\langle x_i\rangle\right)},
\end{equation}
with mean
\begin{equation}
\langle x_i \rangle= e^{-\gamma \Delta t} x_{i-1} + (1-e^{-\gamma \Delta t}) \mu,
\end{equation}
and a covariance matrix given by
\begin{equation}
\Sigma= 2\int_0^{\Delta t}  ds\; e^{\gamma (s-\Delta t)}   \sigma \sigma^T e^{\gamma^T (s-\Delta t)}.
\end{equation}
Both expressions simplify when we set $\sigma = \mathbb{I}$ and $\mu = 0$. For symmetric $\gamma$ ($\gamma=\gamma^T$) we can find an orthogonal eigenbasis $r(\gamma)$ in which the damping matrix is diagonal,
\begin{equation}
\gamma = r(\gamma) \,\Gamma\, r(\gamma)^T,
\end{equation}
where $\Gamma$ is a diagonal matrix and $r(\gamma)r(\gamma)^{T}=\mathbb{I}$. The covariance matrix is also diagonal in this basis, which allows us to compute the integral when $\sigma=\mathbb{I}$ so that
\begin{align}
\Sigma(\Delta t) &= r(\gamma) \,\Gamma^{-1}(1-e^{2\Gamma \Delta t}) r(\gamma)^T.
\end{align}
To ensure that $\gamma$ is symmetric and positive definite (which is required so that the trajectories don't diverge and a steady state exists) we choose the following parametrization:
\begin{equation}
\gamma = \mathbb{I} + \epsilon(d) g,
\end{equation}
where $g$ is a random matrix from the Gaussian Orthogonal Ensemble with density
\begin{equation}
P(g) \propto e^{-\frac{d}{4}\mathrm{Tr}(g^2)}.
\end{equation}
The eigenvalues of $g$ can be both positive and negative, in particular the lowest eigenvalue is distributed according to the Tracy-Widom law with mean $\mu_g=\sqrt{2d}$ and standard deviation of $\sigma_g=\sqrt{2}d^{1/6}$. Choosing $\epsilon(d)=\left(\mu_g+2\sigma_g\right)^{-1}$ ensures that the eigenvalues of $\gamma$ are all positive with good confidence.

\subsection{Neural network architecture and learning hyperparameters} 

{The $M(\phi;\mathcal{I},\mathcal{J})$ objective function is invariant with respect to a global shift in energy,  $M(E^\phi + E_0) = M(E^\phi)$}, since any shift $E_0$ can be incorporated in the partition function $Z$ to obtain the same likelihood-to-evidence ratio. We choose an energy gauge in which the ``free energy'' vanishes, $-\log{Z}=0$. As suggested in~\cite{choi2020regularized} we do so by adding a regularization term of the form $-\lambda_Z (\log Z)^2$ to the likelihood function. Since the constraint $Z=1$ may be satisfied by adding the right constant $E_0$ to the energy function, this regularization does not affect the result of the optimization. We fixed the strength of this term to $\lambda_Z=10^{-3}$.

To perform the benchmark of the methods we used the same neural network architecture for all three objective functions: a multilayer perceptron~\cite{Perceptron0} with two hidden layers of 50 nodes each. 
Each node processes a linear combination of the inputs and adds a constant term (bias). A hyperbolic tangent activation function is then applied to the result of this linear map. 
Between the second hidden layer and the output of the network, we do not apply the activation function.
We implemented $L_2$ regularization on network weights, with regularization strength $\lambda_2$. We optimized the network weights using stochastic gradient descent and the RMSprop~\cite{HINTON} optimization algorithm with learning rate~$l_r$ and size of mini batches~$b$. 

We tuned the hyperparameter by inference of 5 replicate models on $N=10^5$ training data for each objective function and  combination of hyperparameters,  $\lambda_2 \in \{10^{-4}, 10^{-5}, 10^{-6}\}$, $l_r\in \{10^{-2},10^{-3},10^{-4}\}$ and $b\in \{10^{3}, 10^4 \}$. 
We evaluated the mutual information estimate $M(\phi;\mathcal{I},\mathcal{J})$ on $N=10^5$ independent samples (test set). For each of the three methods, we chose hyperparameters for which the mutual information was highest.

\subsection{Methods for likelihood comparison} 

We outline here the method for calculating the Jensen-Shannon divergence between two distributions for which an analytical density is not known but instead we can sample from the two distributions. This will be the case for the likelihood comparison where we will compare the true likelihood and an inferred model of the likelihood.

We generate $M=5 \times 10^4$ samples $\{x^*_m\}_{m=1}^M\sim P(x|\theta^*)$ from the true simulator. In order to generate samples for the inferred estimators  $\{\hat{x}_m\}_{m=1}^M\sim \hat{P}^N_l(x|\theta^*)$ we perform rejection sampling on samples from the marginal probability $P(x)$. To produce samples from $P(x)$ we discard the parameters $\theta$ from the samples $\{(x_i,\theta_i)\}_{i=1}^{N_{\rm tot}}$. Rejection sampling is based on the identity $P(x|\theta)=P(x) Z^{-1} e^{-E (x,\theta)}$ where the likelihood-to-evidence ratio is approximated by an estimator. For each simulation budget $N$ and method $l$ we generate $\{\hat{x}_m\}_{m=1}^M$ samples by rejection sampling with acceptance probability $e^{-E_l^\phi}/Z_l^\phi$. In the last row of Fig.~4 the methods are compared using this metric.

By mixing the samples from $P(x|\theta^*)$ and $\hat{P}^N_l(x|\theta^*)$ in equal proportion we construct an ensemble of samples from $P_{\text{mix}}(x|\theta^*) = \frac{1}{2} (P(x|\theta^*) + \hat{P}^N_l(x|\theta^*))$. We then train two classifiers, one between samples from $P(x|\theta^*)$ and $P_{\text{mix}}(x|\theta^*)$, and the second between samples from $P_{\text{mix}}(x|\theta^*)$ and  $\hat{P}^N_l(x|\theta^*)$. We again exploit the fact that an optimal classifier is the ratio of the two likelihoods and we can read off the two corresponding Kullback-Leibler divergences, $D_{KL}(P||P_{\text{mix}})$ and $D_{KL}(\hat{P}^N_l||P_{\text{mix}})$ from it's estimate (5). The value of the Jensen-Shannon divergence is then the average
\begin{equation}
D_{JS}(P || \hat{P}^N_l) = \frac{1}{2} \left( D_{KL}(P||P_{\text{mix}}) + D_{KL}(\hat{P}^N_l||P_{\text{mix}}) \right).
\end{equation}

An alternative measure of performance of the model of the likelihood is the AUROC characteristic of the optimal classifier between samples from the true and an inferred likelihood, which we will use below. The best models should result in indistinguishable sets of samples for which AUROC$\,=1/2$. A failure to capture the mutual information between parameters $\theta$ and simulated data $x$ make the two sets distinguishable and AUROC$\,=1$.

\subsection{Supplementary benchmark results} 

Fig.~A1 presents the comparison of the objective functions applied to the 4 tasks: the Ornstein-Uhlenbeck process in dimension 1 (A), the birth-death process (B), the SIR process (C), and the Lorenz attractor (D), using the AUROC between the true and estimated likelihood as a measure of performance. The results are consistent with the Jensen-Shannon divergence metric presented in the main text (Fig.~4). 

\begin{figure}[H]
\begin{center}
\includegraphics[width=.9\textwidth]{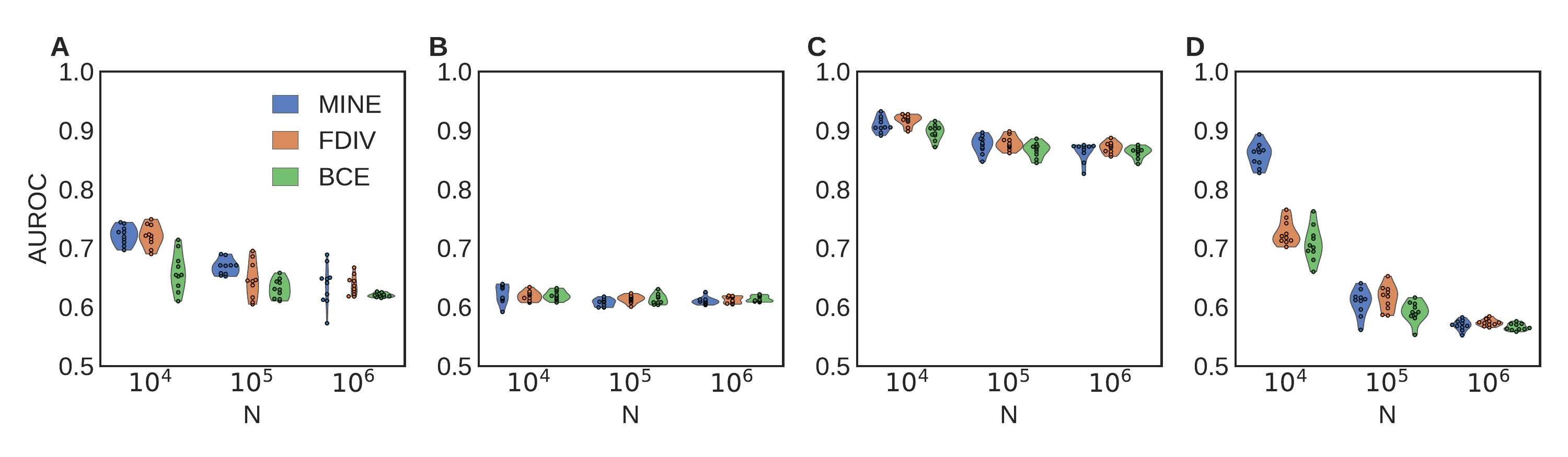}
\caption{We compare the three objective functions $M$ (MINE), $L_f$ (FDIV), and $S$ (BCE) using the Area Under the Receiver Operating Characteristic (AUROC) of a classifier trained to distinguish samples from the simulator $P(x|\theta^*)$ and samples from the inferred estimators $\hat{P}(x|\theta^*)$ for a specific hypothesis $\theta^*$. AUROCs closer to $1/2$ mean that the model is performing well, as its samples are indistinguishable from samples from the true distribution.
This metric gives a consistent picture with respect to the $D_{JS}$ metric presented in the main text (Fig.~4C, F, I, L).} 
\label{illustration}  
\end{center}
\end{figure}

\end{document}